\documentclass{article}
\usepackage{spconf,amsmath,graphicx}
\usepackage[utf8]{inputenc} 
\usepackage{float}
\usepackage{url}

\title{Synthesizing brain tumor images and annotations by combining progressive growing GAN and SPADE}
\name{Mehdi Foroozandeh$^{1}$  \qquad Anders Eklund$^{123}$}

\address{$^{1}$ Division of Medical Informatics, Department of Biomedical Engineering\\
    $^{2}$ Division of Statistics and Machine learning, Department of Computer and Information Science\\
    $^{3}$ Center for Medical Image Science and Visualization\\
    Linköping University, Linköping, Sweden}
\begin{document}
%
\maketitle

\begin{abstract}
Training segmentation networks requires large annotated datasets, but manual annotation is time consuming and costly. We here investigate if the combination of a noise-to-image GAN and an image-to-image GAN can be used to synthesize realistic brain tumor images as well as the corresponding tumor annotations (labels), to substantially increase the number of training images. The noise-to-image GAN is used to synthesize new label images, while the image-to-image GAN generates the corresponding MR image from the label image. Our results indicate that the two GANs can synthesize label images and MR images that look realistic, and that adding synthetic images improves the segmentation performance, although the effect is small. 

\end{abstract}

\vspace{-0.5cm}
\section{Introduction}
\label{sec:intro}

Recent advances in deep learning and computer vision have demonstrated great promise when applied to the automation of tasks that are considered tedious, precise and time-consuming, and that may demand domain-specific expert knowledge to perform. This is relevant not least of all within the field of medical imaging, where such techniques could potentially be leveraged to assist medical experts in tasks and decisions that are especially sensitive and demanding (namely, in patient diagnosis). In the domain of medical image segmentation specifically, it is currently standard practice to annotate medical images such as MR scans by hand; a process which is time-consuming and to some extent subjective (i.e. different medical experts may create different annotations). Thus, any advances towards the successful automation of these tasks would benefit both patients and medical experts greatly, and it is therefore of great interest to investigate if the encouraging results that have been achieved by deep learning and computer vision-based techniques can prove beneficial in this setting. 

However, experimentation is impeded by the fact that medical data is especially scarce and difficult to obtain~\cite{litjens2017survey}, since there are laws (e.g. GDPR) and other security considerations in place that hinder the free distribution of patient data (there are in fact few open neuroimaging datasets available that contain images and segmentations from more than 100 subjects). This is in contrast with other applications within computer vision, such as visual object recognition, which is supported by large, open datasets such as ImageNet~\cite{imagenet} (which contains more than 14 million hand-annotated images). This limitation creates a problem, considering that even the most sophisticated machine learning models only perform as well as the amount of data that they are exposed to.

One attempt at circumventing this issue involves the use of deep learning techniques to generate data \textit{artificially}, in an effort to compensate for the lack of patient data in the medical datasets. This could be achieved by utilizing GANs (Generative Adversarial Networks) --  a deep learning framework that was proposed by Goodfellow et al. in 2014~\cite{goodfellow2014generative} as a method for generating highly realistic data given data sampled from the desired distribution, which has since shown impressive and promising capabilities.

Using synthetic images from GANs to improve training of deep networks is not a new idea. Frid-Adar et al.~\cite{frid2018gan} showed that classical data augmentation (such as adding rotations and making the objects smaller or larger) for liver lesion classification lead to 78.6\% sensitivity and 88.4\% specificity, while adding synthetic images from a 2D GAN improved the performance to 85.7\% sensitivity and 92.4\% specificity. For image segmentation, Bowles et al.~\cite{bowles2018gan} demonstrated that adding synthetic images from a 2D GAN lead to improvements of Dice similarity coefficient between 1 and 5 percentage points. While Bowles et al. synthesized MR images and the corresponding annotations directly, by generating images with two channels, we instead use a two-step process where a noise-to-image GAN~\cite{karras2018progressive} is used to synthesize label images, and an image-to-image GAN~\cite{park2019semantic} is then used to synthesize the MR image from the label image. A similar two-step process was proposed by Guibas et al.~\cite{guibas2017synthetic}. Bowles et al.~\cite{bowles2018gan} worked on binary segmentation of white matter hyperintensities, Guibas et al.~\cite{guibas2017synthetic} worked on retinal fundi images, while we apply our technique to multi-class segmentation of brain tumors. 

\section{Data}

We first introduce the BraTS 2018 dataset~\cite{brats2,bakas2017advancing,bakas2018identifying,bakas2017segmentation1,bakas2017segmentation2} that has been used throughout this project, and explain how the data have been prepared for use in the segmentation network. The Multimodal Brain Tumor Image Segmentation Benchmark (BraTS) is a dataset of volumetric MR scans and corresponding brain tumor segmentations of low- and high grade glioma patients. The MR images in the dataset have been acquired with different clinical protocols and scanners from 19 different institutions, and for each patient 4 types of MR images have been collected:

\begin{itemize}
    \item T1-weighted (T1)
    \item T1-weighted, contrast-enhanced (T1c)
    \item T2-weighted (T2)
    \item T2-weighted FLAIR image (FLAIR)
\end{itemize}

Because of time constraints, the experiments in this project have only been performed using the contrast-enhanced MR images (T1c). The corresponding ground truth segmentation images encompass the following intra-tumoral structures (and the background) as classes:

\begin{enumerate}
    \setcounter{enumi}{-1}
    \item Background (BG)
    \item Necrotic and non-enhancing tumor core (NCR/NET)
    \item Peritumoral edema (ED)
    \item GD-enhancing tumor (ET)
\end{enumerate}

The ground truth segmentations used in this project have been expanded with three more classes in what will be referred to as the \textit{complete} version of the dataset with 7 classes. This version was created by analyzing each subject with the function FAST~\cite{FAST} in the FSL software~\cite{FSL}, with the purpose of obtaining the following new segmentations:

\begin{enumerate}
    \setcounter{enumi}{3}
    \item White matter (WM)
    \item Grey matter (GM)
    \item Cerebrospinal fluid (CSF)
\end{enumerate}

When synthesizing images with GANs, the complete version of the ground truth has been used in favor of the \textit{incomplete} version with 4 classes. This choice is motivated by the expectation that it will be easier for an image-to-image translation network to synthesize complete MR images of the brain from fully segmented images, compared to using tumor segmentations that only cover a small part of the image.

\subsection{Dataset split}
\label{subsection:dataset}

\begin{figure*}[h]
\centering
\includegraphics[width=0.95\textwidth]{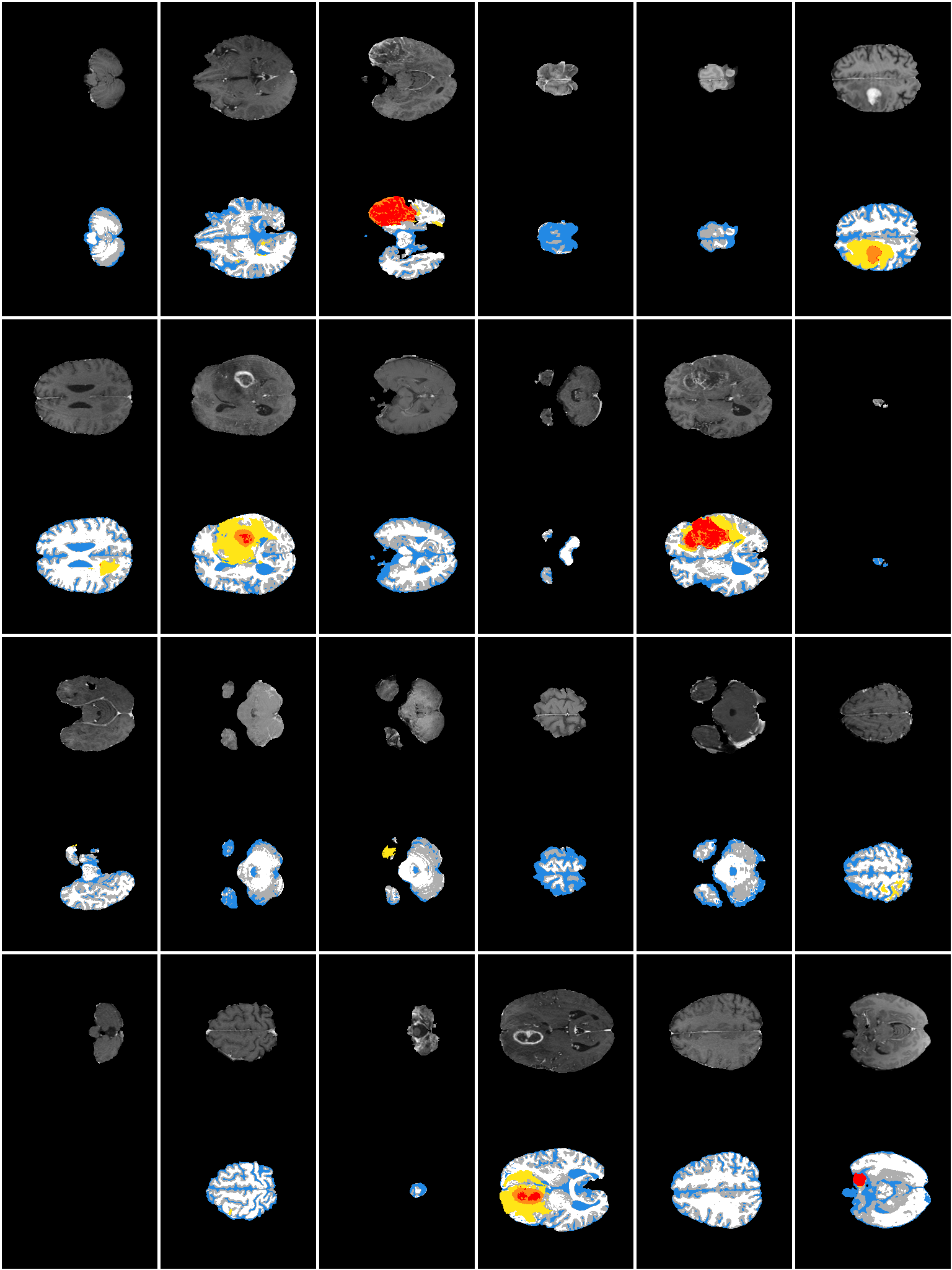}
\caption{24 random (non-empty) slices obtained from the BraTS dataset. The top image in each rectangle shows a T1c MR image, and the bottom image shows its corresponding color-coded segmentation mask. Black: BG, white: WM, grey: GM, blue: CSF, yellow: ED, orange: ET, red: NCR/NET.}
\label{fig:brats}
\end{figure*}

The BraTS 2018 dataset used in this project consists of 210 high grade glioma pairs of 3D MR volumes and corresponding annotations, represented in the NIFTI~\cite{nifti} file format. Each volume contains $240 \times 240 \times 155$ voxels with an isotropic voxel size of $1 \times 1 \times 1$ mm. These \texttt{.nii}-files were read into Python with the  Nibabel~\cite{nibabel} library and sliced axially (i.e. in a direction parallell with a line going from the chin to the top of the head) into 155 2D slices each, resulting in a total number of $210 \times 155 = 32,550$ 2D slices. The slicing was performed using Numpy~\cite{numpy}, a Python library for matrix and array calculations, and the resulting slices were saved \textit{without loss or corruption of array data} into separate \texttt{.png} files using the Python library PyPNG~\cite{pypng}; the MR images (which store 16 bit information) were saved in a greyscale \texttt{uint16} file format and the segmentation masks (which are simply integer-valued matrices where each integer represents a class) were saved in a greyscale \texttt{uint8} file format. Before being saved as image files, each slice was padded with zeroes around the border to round up the resolution to the nearest power of two, i.e. to $256 \times 256$ (a requirement by PGAN). The resulting 32,550 \texttt{.png} files were subsequently shuffled, and separated into training (80 \%), validation (10 \%) and test data (10 \%). See figure~\ref{fig:brats} for an image of 24 random samples from the training set.

\section{Methods}
\label{sec:format}

We will here detail the steps that have been taken to investigate the idea outlined in the introduction, i.e. the process of training a segmentation network on a dataset of brain MR images, synthesizing new brain images with GANs and including these in the segmentation. We start with a section on segmentation, which will provide details surrounding the application of U-Net and the architecture and hyperparameters that have been employed. Following this is two sections on progressive GAN (PGAN)~\cite{karras2018progressive} and SPADE respectively~\cite{park2019semantic}, which will explain how these GAN frameworks have been applied in this project. Lastly, there will be a section that describes the preprocessing operations that have been applied on the synthetic segmentation masks.

\subsection{Segmentation}
\label{section:segmentation}

The segmentation tasks in this project use an implementation of U-Net~\cite{unet} in the Python deep learning library Keras (run on top of the machine learning platform TensorFlow). The network architecture used here is a rather standard U-Net. To reiterate: the encoder part consists of four levels with two convolutions per level, with 64, 128, 256 and 512 convolutional filters per convolution in each respective level. Each consecutive encoder level is connected by a max pooling layer. The fifth level is the ''bridge'' between the encoder and decoder part, which consists of a convolution with 1024 filters followed by a transposed convolution. The decoder part that follows consist of four levels with two convolutions each with 512, 256, 128 and 64 filters per convolution in each respective level, and each consecutive level is connected by a transposed convolution layer. The last convolution is followed by a final convolution layer where the number of filters is the number of classes in the given dataset. Each convolution layer uses 'same' padding, and is followed by a batch normalization layer. Each transposed convolution layer uses a $3 \times 3$ kernel with 'same' padding and a $(2,2)$ stride, and is followed by a batch normalization layer (before concatenation). The final convolution layer is followed by a softmax activation function applied over the channel axis, which results in a multi-channel segmentation map in which the value in a given channel corresponds to the probability that the pixel in question belongs to the class represented by that channel (a single-channel image can then be created post-training by taking the argmax of the segmentation map over the channel axis). The respective weights of the convolution and transposed convolution kernels are initialized with He normal initialization~\cite{henormal}.

Before being fed to the network, the MR images in the training, validation or test sets were scaled and normalized with respect to the training set. This was done by dividing each pixel value with a constant equal to the maximum pixel value in the training set, followed by subtracting the scalar mean of all (scaled) pixel values in the training set. Furthermore, each segmentation mask was converted to a multi-channel one-hot encoded tensor before entering the segmentation network.

The training was performed via stochastic gradient descent with the Adam~\cite{adam} optimization algorithm, using a learning rate of $10^{-4}$ and a batch size of 8. The training and validation sets were read sequentially and shuffled after each respective epoch (i.e. after all of the images in each respective dataset had been used once). Weighted Dice loss was used as the loss function and the validation metric, and the corresponding weight vector was calculated over the training set.

Using these settings, different training instances were created with varying numbers of real and synthetic training data, resulting in a different training set per instance (however, the validation and test data remained identical between all instances). The number of segmentation classes were varied as well; apart from the complete version with 7 classes, the network was also trained with the incomplete version of the dataset (4 classes) as well as with a binary version (2 classes: tumor or non-tumor). The datasets with less than 7 classes were either read directly from disk when available, or created from a more complete dataset at training time by setting all irrelevant classes to 0 and the tumor classes to values in $[1,2,3]$ (in case of a 4-class problem), or $1$ (in case of a binary problem).

Each training instance was run for 150 epochs, and the network weights were saved every time a lower validation error was achieved at the end of an epoch. The results are based on the weight configurations that have been achieved after these 150 epochs (i.e. the weights that yield the lowest validation error). The training was performed using an NVIDIA RTX 2080 Ti and two NVIDIA GTX 1080 GPUs (separately). See the results section for the complete list of training instances.

\subsection{PGAN}
\label{section:PGAN_method}

The official implementation of PGAN~\cite{karras2018progressive}\footnote{https://github.com/tkarras/progressive\_growing\_of\_gans} was downloaded and trained on the complete (7-class) dataset of segmentation masks from BraTS to generate a new, synthetic dataset of segmentation masks. Default settings were used, apart from changing the '\texttt{dynamic\_range}' parameter in the configuration file from $[0, 255]$ to $[0, 6]$, to ensure that the transformations between the network values (which are continuous and lie in $[-1, 1]$) and the values of the segmentation masks (which assume discrete values between 0 and 6) were performed correctly. Additionally, slight modifications were made to the scripts responsible for generating and saving image files (namely, \texttt{'util\_scripts.py'} and \texttt{'misc.py'}), in order to ensure that the final generated outputs were saved as \texttt{.png} files with the same properties as the ones in section \ref{subsection:dataset}. Similar changes were also made to the scripts related to the calculation of the image metrics.

Two training instances were created: one that used the ''full'' dataset consisting of 100 \% of the training data (26,040 images), and one that used a ''reduced'' dataset consisting of only the first 20 \% of the images in the full dataset (5,208 images). Completely empty segmentation masks (which comprised 15.66 \% of the full dataset and 15.44 \% of the reduced dataset) were discarded when loaded into the scripts, resulting in 21,962 and 4,404 training images respectively. The training script was set to run until $12 \cdot 10^{6}$ of the training images had  been used in the training loop (as dictated by the parameter '\texttt{total\_kimg}', which by default is set to 12,000).

In both training instances, the network weights were saved after every 'tick' (or iteration) of the training loop. After the training, the SWD score was calculated over the generated images with respect to each of the saved weights, and the weights that yielded the lowest (average) distance were saved and used to generate a dataset of 100,000 images -- this was done once per training instance, resulting in two new datasets of segmentation masks. Both training instances were run using an NVIDIA RTX 2080 Ti GPU (at separate instances).

\subsection{SPADE}

The official implementation of SPADE~\cite{park2019semantic}\footnote{https://github.com/NVlabs/SPADE} was downloaded and trained on the 7-class version of the BraTS dataset to learn a ''mapping'' from segmentation masks to an MR image represenation. The synthetic segmentation masks generated in section \ref{section:PGAN_method} were subsequently given to the trained network to generate their MR counterparts.

Similiarly to in section \ref{section:PGAN_method}, the source code had to be modified slightly in order to adapt it to the properties of the dataset. In particular, the data loading and utilities scripts (namely, '\texttt{pix2pix\_dataset.py}' and '\texttt{util.py}') had to be altered at places to accommodate single-channel MR images with values larger than 255, and to ensure that the transformations between tensor and image values were performed correctly. Again, small changes also had to be made (specifically to '\texttt{util.py}')  in order to ensure that the generated MR images were saved as \texttt{.png}-files with the desired properties.

Default settings were used, apart from disabling image preprocessing, image flipping, inclusion of instance maps and VGG loss calculation (via the input argument \texttt{no\_vgg\_loss}). Like in section \ref{section:PGAN_method}, two training instances were created: one with 100 \% of the training data, and one with the first 20 \% of the training data. The instance with 100 \% of the training data was trained for the default number of 50 epochs, and the instance with 20 \% training data was trained for 250 epochs. The reduced instance was trained for five times as many epochs as the full dataset, in order to ensure that the network would be shown the same number of training data in both instances (like in PGAN), and since this yielded better image quality, subjectively speaking.

After each training instance, the segmentation masks generated in section \ref{section:PGAN_method} were preprocessed (see section \ref{section:preprocessing}) and the resulting images were used with the final weights of the generator to create their MR counterparts. The segmentation masks that were created with 100 \% of the training data were input to the SPADE generator trained with 100 \% of the data, and the masks that were generated with 20 \% of the training data were used with the SPADE generator that was trained with the same 20 \% of the data, ultimately resulting in two datasets of synthetic segmentation masks and corresponding MR images. Each training instance was performed using two NVIDIA Tesla V100 simultaneously (at separate instances).

\subsection{Preprocessing}
\label{section:preprocessing}

A simple preprocessing procedure was applied to the segmentation masks generated in section \ref{section:PGAN_method}, in a quick effort to remove noisy or corrupted images from the synthetic datasets. This was done by comparing every image in a synthetic dataset to the entirety of the (corresponding) real dataset, by calculating the \textit{Z-score} of each pixel value in the synthetic dataset with respect to the real dataset; i.e. by subtracting the empirical mean of the pixel values in the real dataset from each pixel value in the synthetic dataset, and subsequently diving them by the standard deviation of each pixel value in the real dataset. The mean and standard deviation was calculated over the \textit{batch} axis of the real dataset, resulting in two two-dimensional arrays of values in which each pixel value represents the mean or standard deviation of a pixel value at a given \textit{location} in the segmentation map. The resulting \textit{standardized} images provide a measure of how much each pixel in a specific location in an image of the synthetic dataset deviate from all of the pixels in the same location, in the real dataset.

Following this, each image in the standardized synthetic dataset was reshaped into a vector, and the Euclidean norm of each vectorized image was calculated, resulting in a new vector where each value corresponds to an image in the synthetic dataset. These values were created with the intent of providing a measure of how much each synthetic image deviates from the entirety of the real dataset, and thus a \textit{threshold} value was applied to every element in this vector to determine if the corresponding synthetic image should be kept or discarded; if a particular norm was larger than this threshold value, the corresponding image was discarded, otherwise it was saved.

The value of the threshold was chosen by experimentation with the intent of maximizing the fraction of synthetic images that were discarded when applying the threshold, while minimizing the fraction of real images that would have been discarded if the same procedure would have been applied to the real dataset. The threshold value of 500 was chosen to preprocess both datasets of synthetic segmentation masks, i.e. the ones synthesized using 100 \% and 20 \% of the real data respectively. Both of these datasets used those respective fractions of real data as reference when calculating their respective Z-scores. See section \ref{section:preprocessing_results} for image examples of segmentation masks that were discarded.

\section{Results}
\label{sec:typestyle}

We will start with results from the PGAN; image examples, training duration and miscellaneous information will be presented. This will be followed by a section on the preprocessing steps covered in section \ref{section:preprocessing}, which will mainly consist of examples of images that have been discarded from the final synthetic datasets via this process. Following this is a section on the MR images that have been generated by SPADE, where image examples and training duration will be presented. Lastly, a section on segmentation will follow which will include test scores from all of the training instances that were created via the methods described in section \ref{section:segmentation}, as well as image examples of the corresponding segmentation results. Apart from discarding completely empty images, \textit{none of the image examples in this chapter have been cherry-picked}.

\subsection{PGAN}

In this section, the generated image examples, training duration and distance metrics will be presented for the PGAN instances that were trained on the full and reduced datasets respectively

\subsubsection{Full dataset}

\begin{figure*}[h]
\centering
\includegraphics[width=0.95\textwidth]{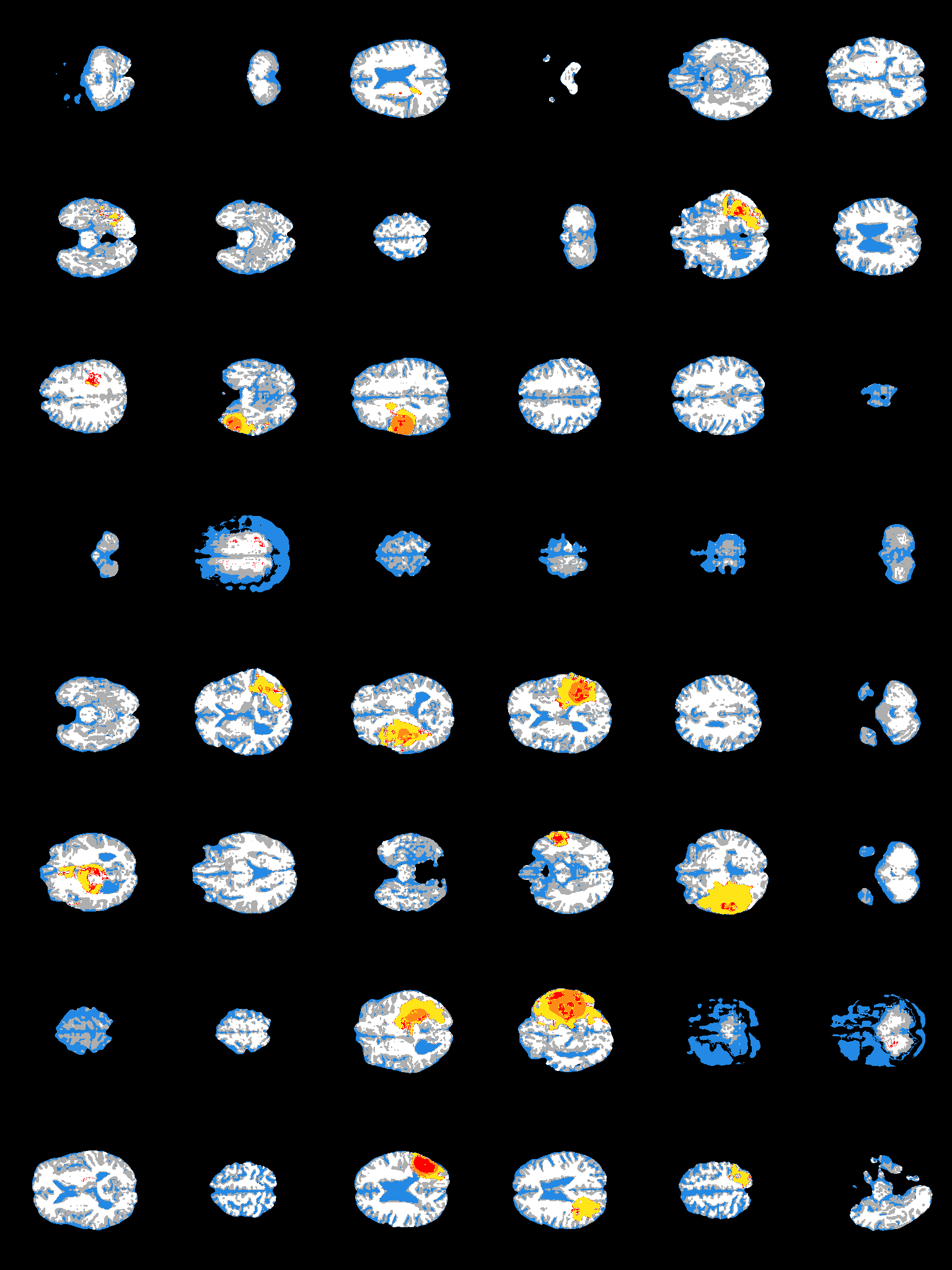}
\caption{48 examples of segmentation masks generated by the instance of PGAN that has been trained on the full dataset.}
\label{fig:PGAN_results_full}
\end{figure*}

The training instance that used the full dataset (21,962 non-empty 7-class segmentation masks) required 6 days, 10 hours and 46 minutes of training to reach a total showing of $12 \cdot 10^{6}$ images, or 205 iterations. The weight configuration that yielded the lowest average SWD score of 3.2130 was achieved after 1 day, 19 hours and 59 minutes, which corresponds to $7.2401 \cdot 10^{6}$ images, or 86 iterations. See figure \ref{fig:PGAN_results_full} for 48 examples of (unprocessed) segmentation mask that were generated with these weights.

\subsubsection{Reduced dataset}

\begin{figure*}[h]
\centering
\includegraphics[width=0.95\textwidth]{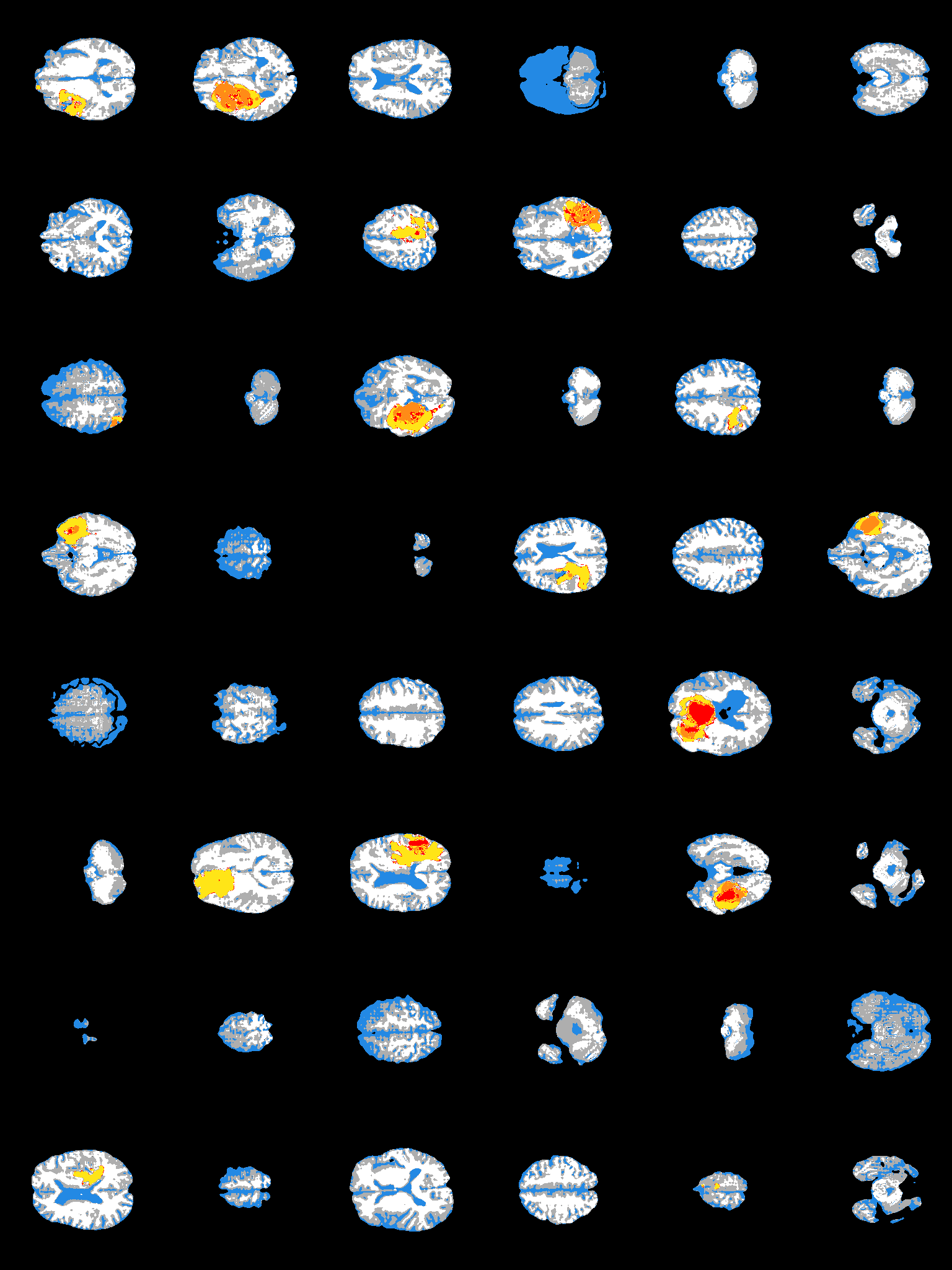}
\caption{48 examples of segmentation masks generated by the instance of PGAN that has been trained on the reduced dataset.}
\label{fig:PGAN_results_fifth}
\end{figure*}

The training instance that used the reduced dataset (4,404 non-empty 7-class segmentation masks) required 6 days, 11 hours and 0 minutes of training to reach a total showing of $12 \cdot 10^{6}$ images, or 205 iterations. The weight configuration that yielded the lowest average SWD score of 3.5915 was achieved after 1 day, 22 hours and 58 minutes, which corresponds to $7.3601 \cdot 10^{6}$ images, or 89 iterations. See figure \ref{fig:PGAN_results_fifth} for 48 examples of (unprocessed) segmentation mask that were generated with these weights.

\subsection{Preprocessing}
\label{section:preprocessing_results}

\begin{figure*}[h]
\centering
\includegraphics[width=0.95\textwidth]{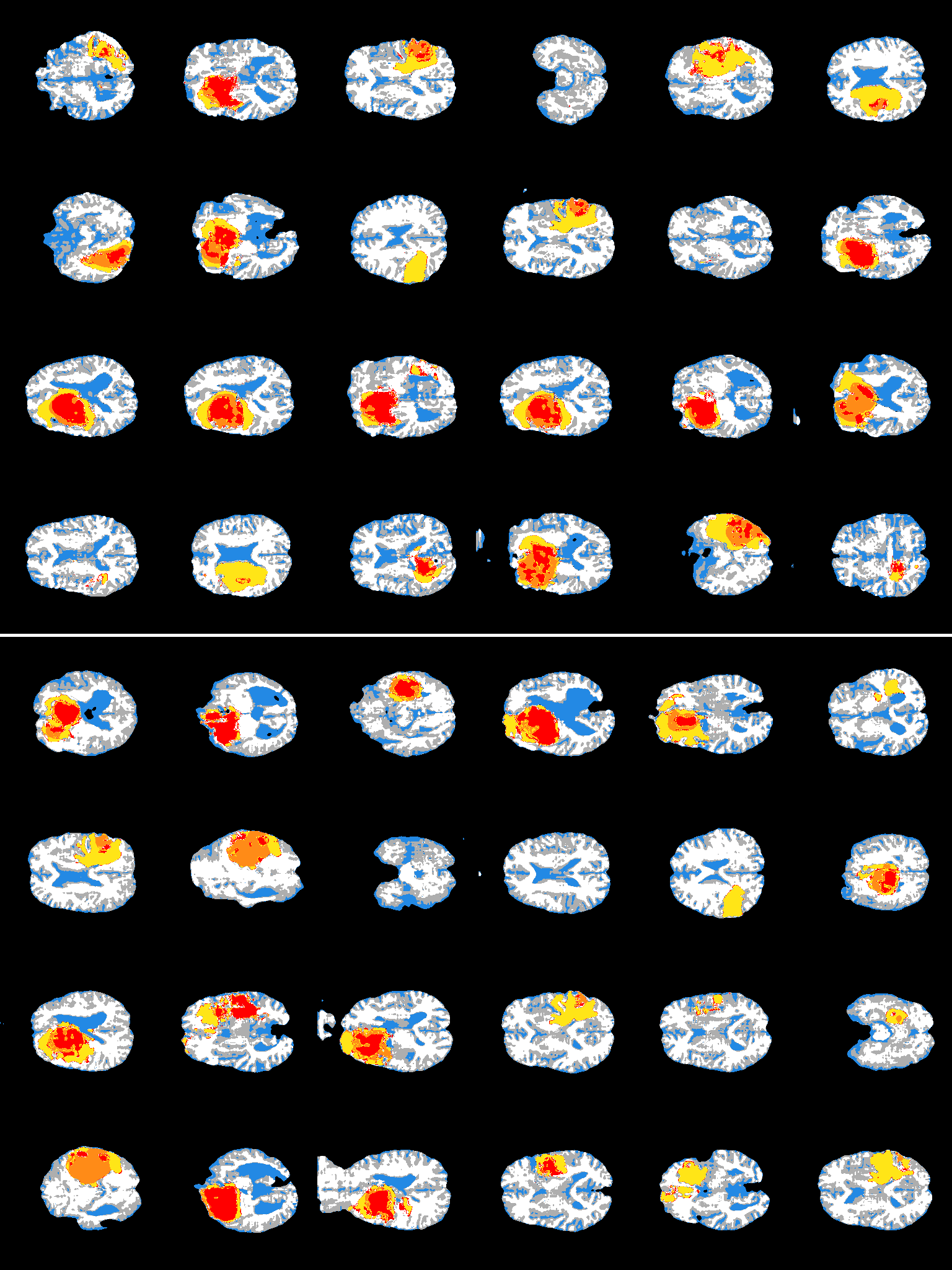}
\caption{24 examples of images that have been discarded via preprocessing from the full and reduced dataset of PGAN-generated segmentation masks respectively. Top: full dataset, bottom: reduced dataset.}
\label{fig:preprocessed}
\end{figure*}

The application of the steps described in section \ref{section:PGAN_method} on the full and reduced datasets removed 0.597 \% and 1.882 \% of the segmentation masks respectively. This method was not successful at removing the ''blurry'' masks that were relatively common in both datasets, but it managed to remove the small fraction of masks with artefacts in the shape of image patches in inappropriate locations. See figure \ref{fig:preprocessed} for 24 examples of the segmentation masks that have been discarded from the respective datasets.

\subsection{SPADE}

\begin{figure*}[h]
\centering
\includegraphics[width=0.95\textwidth]{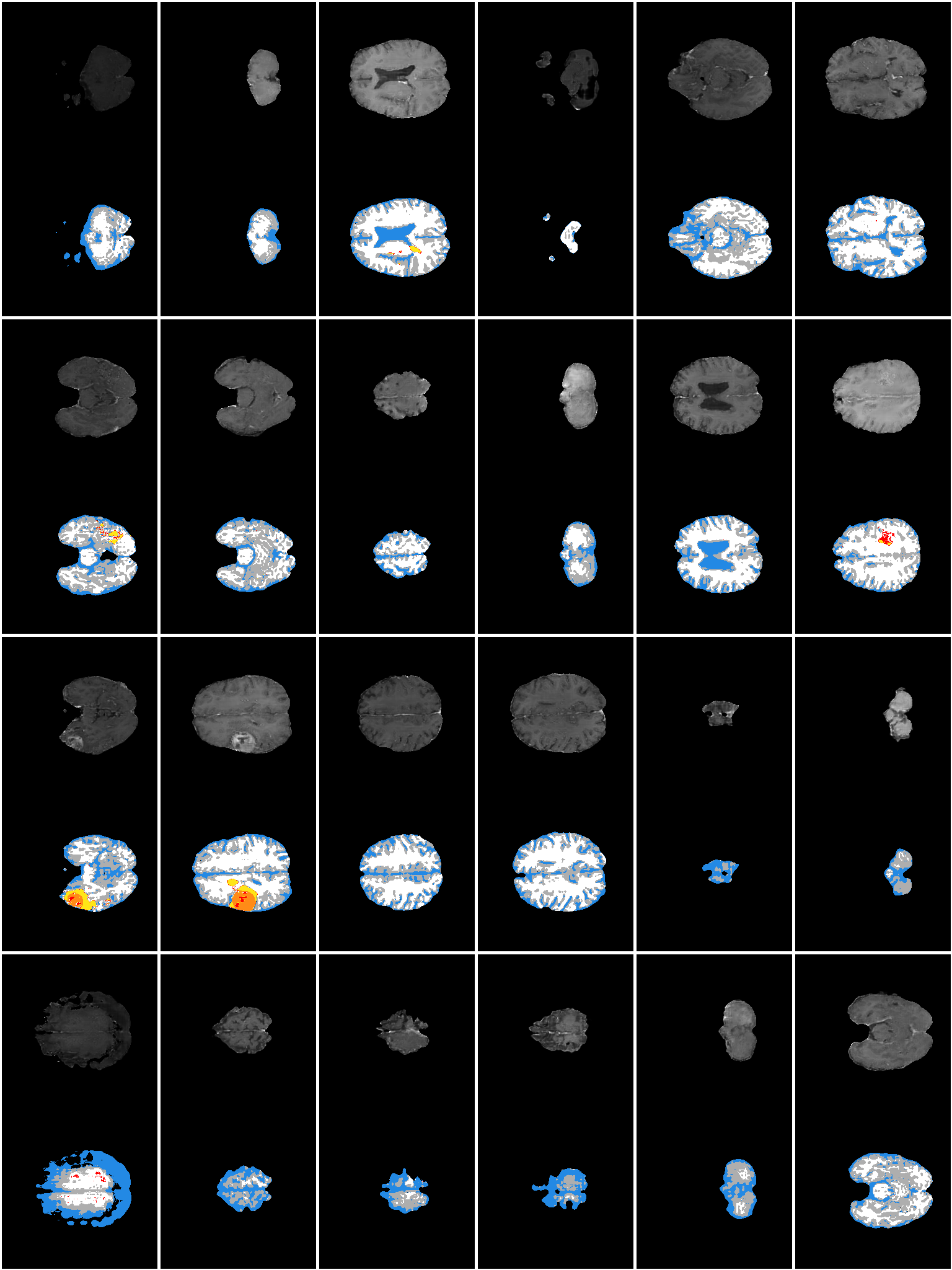}
\caption{24 examples of pairs of synthetic MR images and corresponding synthetic segmentation masks, generated from the full BraTS dataset. The top image in each rectangle shows an MR image generated by SPADE, and the bottom image shows the corresponding PGAN-generated segmentation mask that was used to generate it.}
\label{fig:spade_full}
\end{figure*}

\begin{figure*}[h]
\centering
\includegraphics[width=0.95\textwidth]{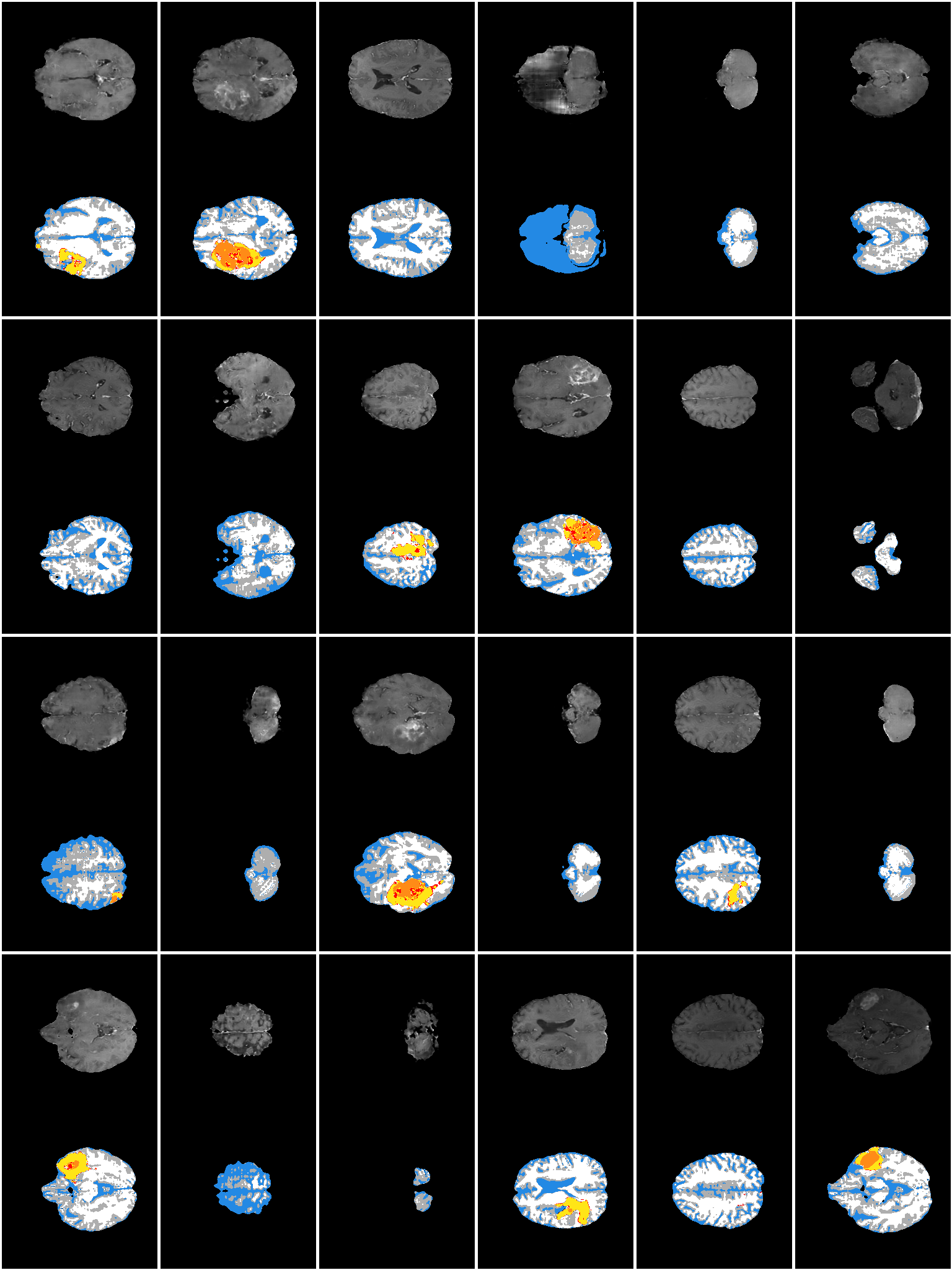}
\caption{24 examples of pairs of synthetic MR images and corresponding synthetic segmentation masks, generated from the reduced BraTS dataset. The top image in each rectangle shows an MR image generated by SPADE, and the bottom image shows the corresponding PGAN-generated segmentation mask that was used to generate it.}
\label{fig:spade_fifth}
\end{figure*}

\begin{figure*}[h]
\centering
\includegraphics[width=0.85\textwidth]{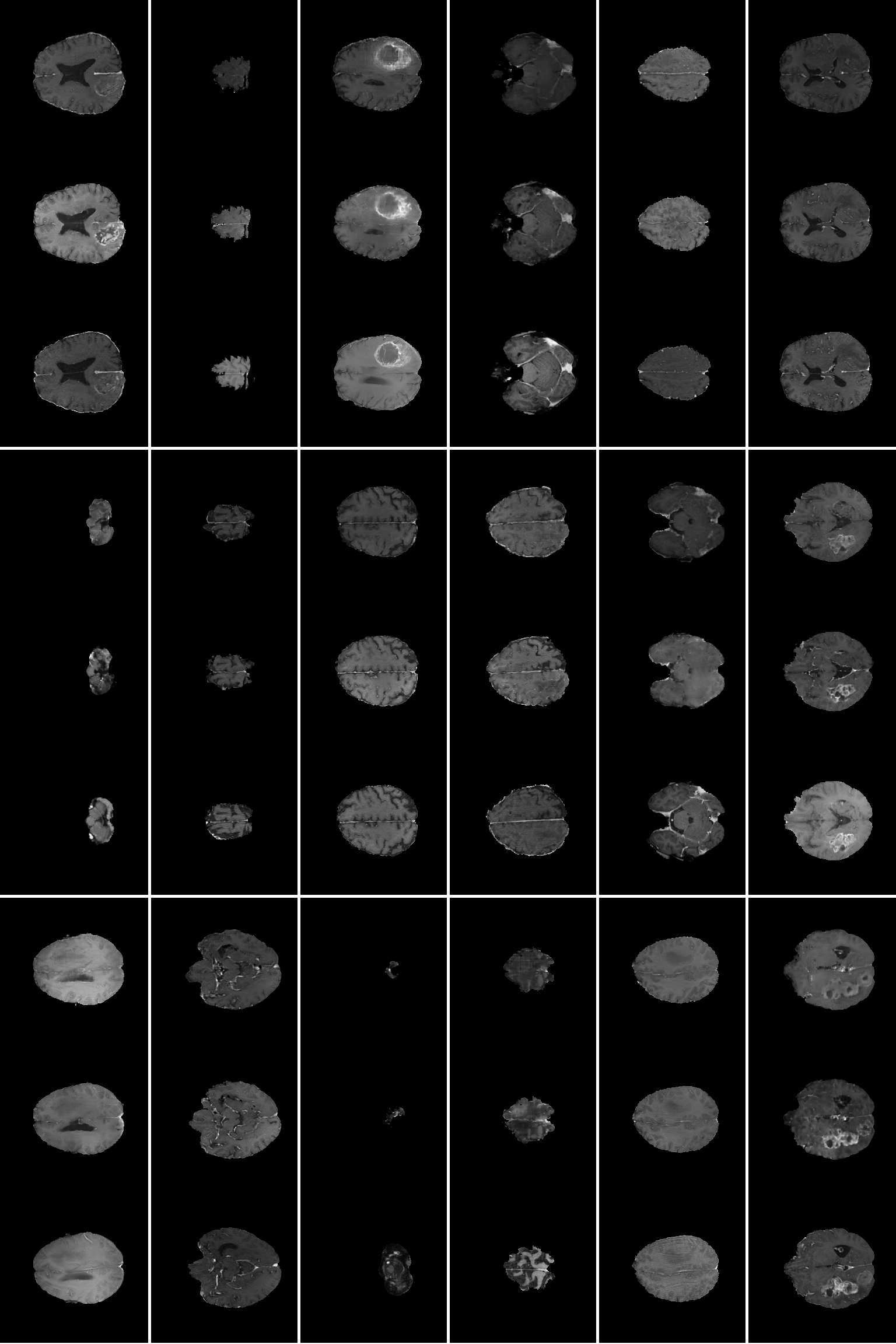}

\caption{18 examples of MR images synthesized from both (full and reduced) training instances, when given (real) segmentation masks from the test set as inputs. The top and middle images in each rectangle shows MR images that have been generated by the full and reduced SPADE instances respectively, and the image below them shows the corresponding \textit{true} MR image from the test set.}

\label{fig:spade_test}
\end{figure*}

The SPADE instance that was trained on the full BraTS dataset (26,040 image pairs) required 2 days, 13 hours and 26 minutes to complete a total of 50 epochs. The instance that was trained on the reduced dataset (5,208 image pairs) required 3 days, 7 hours and 2 minutes to complete a total of 250 epochs. See figure \ref{fig:spade_full} and \ref{fig:spade_fifth} for 24 examples respectively of MR images generated with the preprocessed segmentation masks generated via PGAN. Additionally, see figure \ref{fig:spade_test} for examples of MR images generated with the (real) test data from BraTS.

\subsection{Segmentation}
\label{results:segmentation}

\begin{figure*}[h]
\centering
\includegraphics[width=0.95\textwidth]{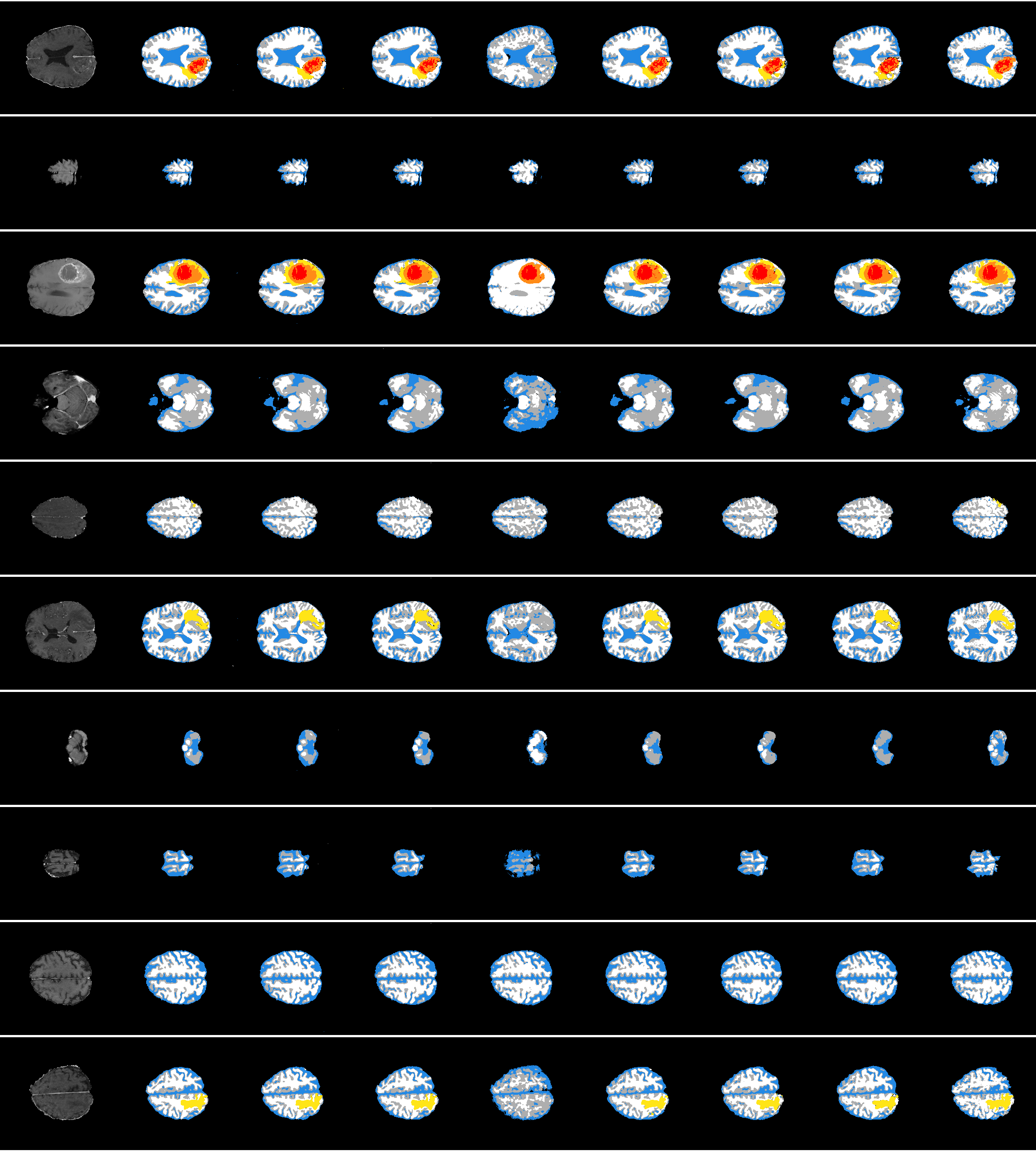}

\caption{10 examples of 7-class segmentations of MR images from the test set, created using the final results of the different training instances. The first image in each row is the input MR image, and the last image is the ground truth segmentation. The images in-between are segmentation results created with the weights of the different training instances, ordered by the rows in table \ref{table:dice}; i.e. the first segmentation corresponds to the training instance created with 26,040 real and 0 synthetic images, the second with 26,040 real and 8,960 synthetic images, and so on.}

\label{fig:segmentations_7}
\end{figure*}

\begin{figure*}[h]
\centering
\includegraphics[width=0.9\textwidth]{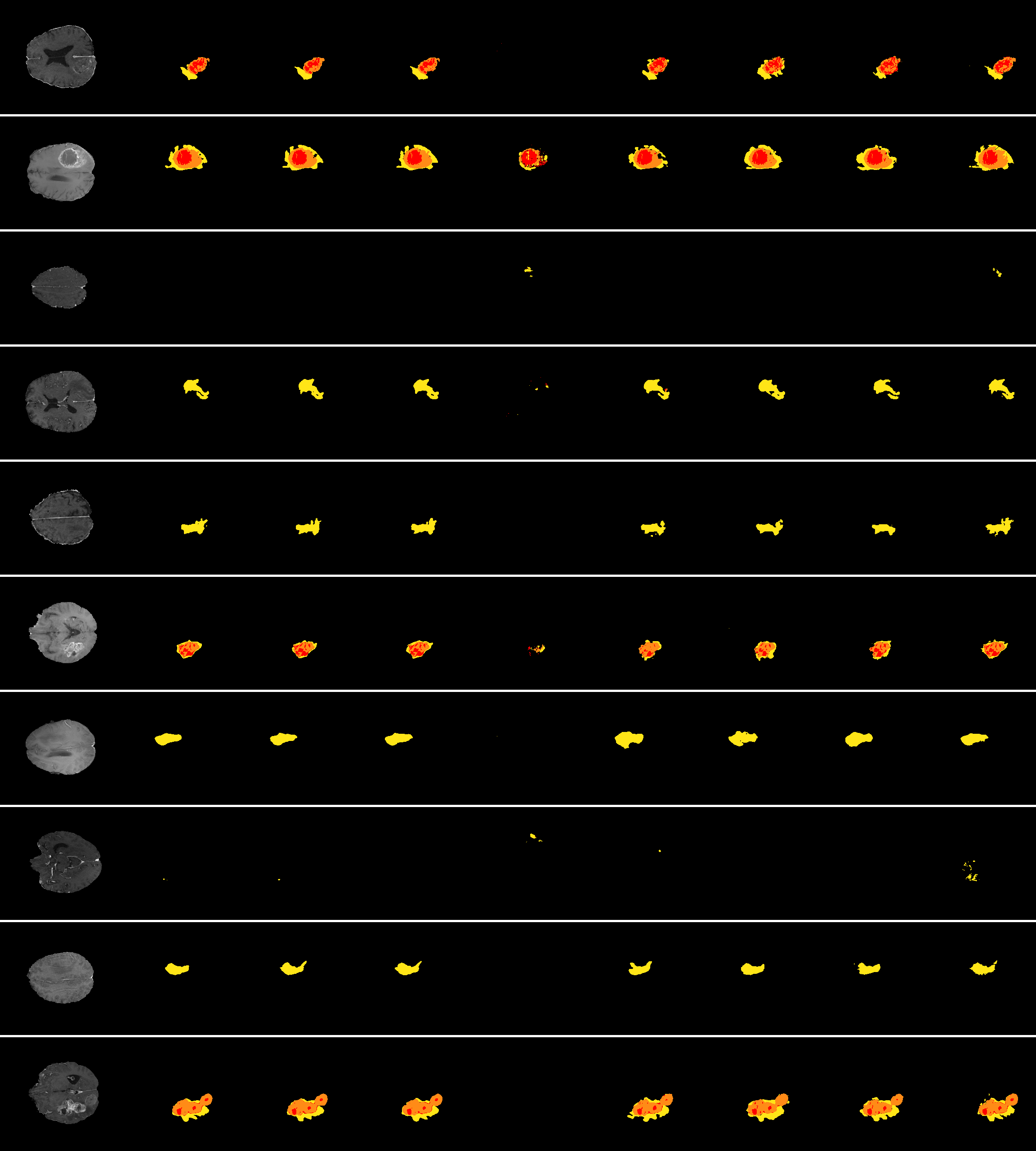}

\caption{10 examples of 4-class segmentations of MR images from the test set, created using the final results of the different training instances. The first image in each row is the input MR image, and the last image is the ground truth segmentation. The images in-between are segmentation results created with the weights of the different training instances, ordered by the rows in table \ref{table:dice}; i.e. the first segmentation corresponds to the training instance created with 26,040 real and 0 synthetic images, the second with 26,040 real and 8,960 synthetic images, and so on.}

\label{fig:segmentations_4}
\end{figure*}

\begin{figure*}[h]
\centering
\includegraphics[width=0.9\textwidth]{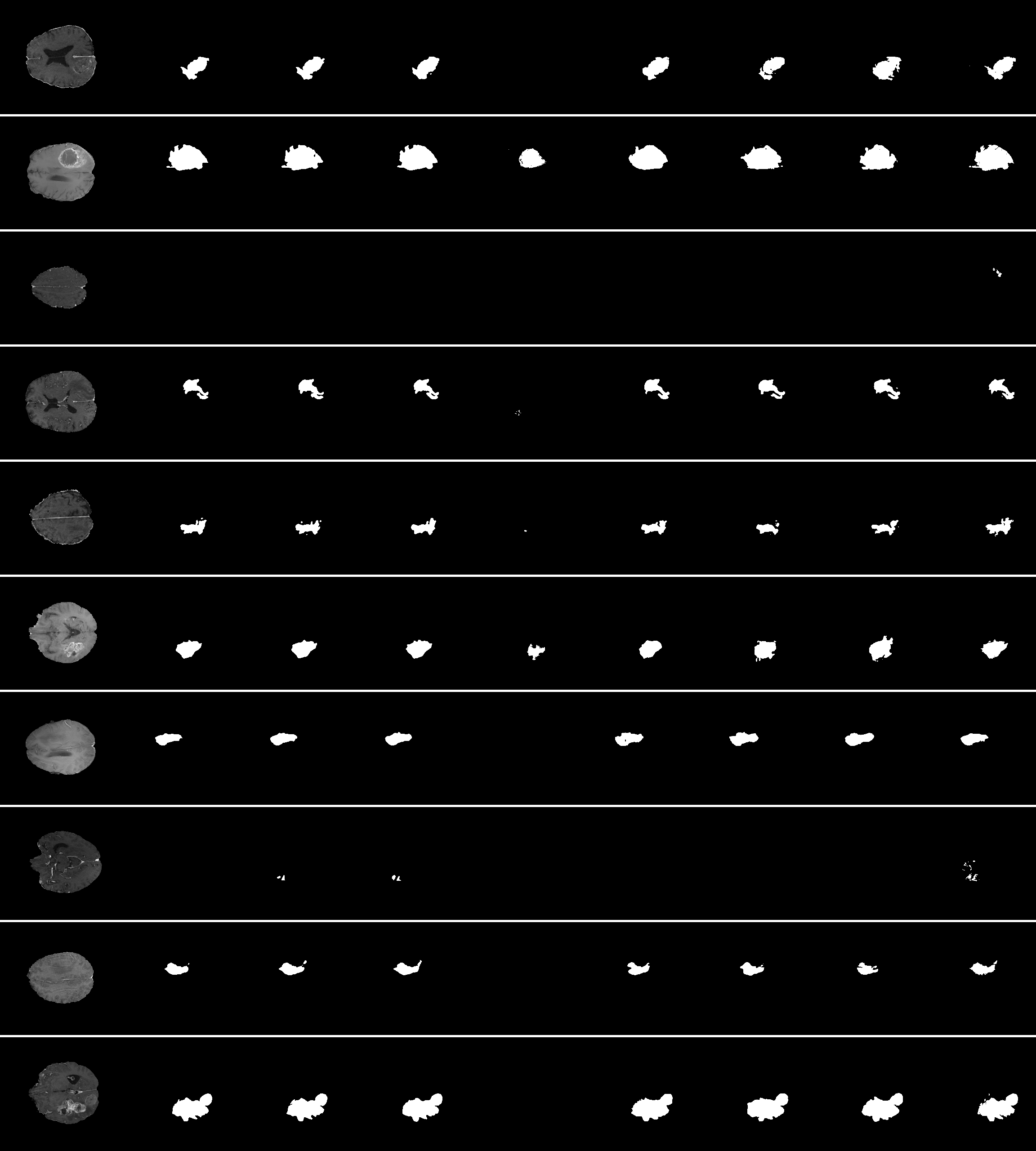}

\caption{10 examples of binary segmentations o MR images from the test set, created using the final results of the different training instances. The first image in each row is the input MR image, and the last image is the ground truth segmentation. The images in-between are segmentation results created with the weights of the different training instances, ordered by the rows in table \ref{table:dice}; i.e. the first segmentation corresponds to the training instance created with 26,040 real and 0 synthetic images, the second with 26,040 real and 8,960 synthetic images, and so on.}

\label{fig:segmentations_2}
\end{figure*}

Depending on the GPU that was used, the number of data and the number of classes (problems with fewer classes required less training time), each segmentation instance required between 0.65 to 3.7 full days of training to complete 150 epochs. The weighted Dice loss was calculated over the test set using the latest saved weights of each respective instance. The results are presented in table \ref{table:dice}, and image examples of corresponding segmentation results are shown in figure \ref{fig:segmentations_7}, \ref{fig:segmentations_4} and \ref{fig:segmentations_2}.

\setlength{\tabcolsep}{20pt}
\renewcommand{\arraystretch}{1.5}

\begin{table*}[t]
\centering
\resizebox{0.8\linewidth}{!}{%
    \begin{tabular}{l|l|l|l|}
    \cline{2-4}
                                                                               & \multicolumn{3}{c|}{\textbf{Classes}}                                    \\ \hline
    \multicolumn{1}{|c|}{\textbf{Number of training images (real, synthetic)}}                      & \multicolumn{1}{c|}{7} & \multicolumn{1}{c|}{4} & \multicolumn{1}{c|}{2} \\ \hline
    \multicolumn{1}{|l|}{(26,040, 0), total: 26,040}     & 10.94                  & 6.62                   & 2.49                   \\ \hline
    \multicolumn{1}{|l|}{(26,040, 8,960), total: 35,000}  & 10.92                  & 6.65                   & \textbf{2.48}          \\ \hline
    \multicolumn{1}{|l|}{(26,040, 23,960), total: 50,000} & \textbf{10.90}         & \textbf{6.42}          & 2.53                   \\ \hline
    \multicolumn{1}{|l|}{(0, 26,040), total: 26,040}     & 42.76                  & 39.11                  & 24.14                  \\ \hline
    \multicolumn{1}{|l|}{(5,208, 0), total: 5,208}       & 16.80                  & 12.90                  & 6.16                   \\ \hline
    \multicolumn{1}{|l|}{(5,208, 4792), total: 10,000}   & 16.44                  & 12.51                  & 6.14                   \\ \hline
    \multicolumn{1}{|l|}{(5,208, 20832), total: 26,040}  & \textbf{16.18}         & \textbf{12.11}         & \textbf{5.80}          \\ \hline
    \end{tabular}
    }
    \caption{Dice error in percent, calculated over the test set. \textbf{Top rows:} training with the full dataset (26,040 images). \textbf{Bottom rows:} training with the reduced dataset (5,208 images). The best data configuration in each class has been highlighted in bold, with respect to both the full and reduced datasets.}
    \label{table:dice}
\end{table*}

Interestingly, the instances that were trained using synthetic images exclusively (which were created using the \textit{full} training set) displayed very poor training progress; the training instance using the complete (7-class) synthetic dataset was particularly notorious since its weights only got updated once and in the first epoch (that is, the following 149 epochs saw \textit{no} improvements in validation error). The corresponding 4-class instance saw only 5 improvements in the first 51 epochs, and the binary instance saw 4 improvements in the first 32 epochs.

\section{Discussion}
\label{sec:typestyle}

The inclusion of synthetic data slightly improved test scores across all class subsets and training set fractions, depending on the quantity used -- some quantities were detrimental, but at least one of the quantities in each category reduced the Dice error compared to the corresponding training instance that only used real data. The benefits were strongest in experiments involving only a fifth of the training data, and with the exception of the binary training instances using the full dataset, all of the training instances showed a preference for the data quantitites with the largest amount of synthetic images. This seems to support the idea that data augmentation using GANs is worth considering when one is restricted to using small datasets, and these results are aligned with those by Bowles et al.~\cite{bowles2018gan}. As stated however, these improvements are modest, and the visual quality of the synthetic segmentation masks generated by PGAN is mixed (although the mapping from segmentation masks to MR images appears to be more consistent -- see figure \ref{fig:spade_test}). Some of the generated segmentation masks look convincing, while many appear corrupt and noisy. No traditional augmentation (e.g. rotations) was used when training PGAN and SPADE, this could potentially improve the results.

In summary, these results are encouraging, but somewhat preliminary. More experimentation is necessary in order to improve the quality of the segmentation masks, and it would be of interest to observe how much more significant the observed benefits would be if the same experiments were performed using synthetic datasets of higher visual quality than the ones generated in this project. 

The synthetic segmentation masks that have been generated via PGAN look visually convincing at first glance, but appear somewhat noisy upon closer inspection; especially with respect to the tumoral classes present in the ''incomplete'' version of the BraTS dataset, i.e. class 1, 2 and 3  (red, yellow and orange), which are obviously the most important classes with regards to the segmentation of tumors. Furthermore, a significant portion of the generated image masks appear corrupt and heavily noisy, even after the somewhat superficial attempt at image preprocessing. This might stem from or be affected by the large variation in the \textit{size} of the slices in the real dataset, which contains slices with sizes ranging from just a few pixels to the majority of the $256 \times 256$ image area. This is possibly making the job of the generator more difficult since it is forced to model a probability distribution that has to capture variation over both size and anatomy (which, by not being independent, further adds to the complexity of the probability distribution that is to be modelled). Following this intuition, it may be of interest to generate a synthetic dataset of segmentation masks by creating multiple training instances of PGAN on separate subsets of the real dataset, created by slicing each brain in a smaller range of the slicing axis (e.g. to create 31 training instances of 1050 images, where each image is in a separate 5-pixel range of the axial axis, whose length is 155 pixels). This would allow each separate PGAN instance to become specialized in learning a distribution over a specific range of slices, instead of the entire three-dimensional brain. Another idea is to instead use 3D GANs~\cite{eklund2019feeding,abramian2019generating,cirillo2020vox2vox}.

Interestingly, the visual quality of the synthetic dataset of segmentation masks that was generated using a fifth of the training data does not appear to differ significantly from the dataset generated using the full BraTS dataset, and the difference in SWD score of 0.3785 is perhaps smaller than one might expect. This might have been echoed by the fact that the segmentation instances that used the reduced dataset saw a significantly greater boost in performance when mixed with synthetic data, compared to the instances that did the same but with the full dataset of real images, where the effects on performance were minuscule and sometimes even detrimental (as seen in (2; 26,040, 23,960) and (4; 26,040, 8,960) in table \ref{table:dice}). However, it is possible that this has more to do with the U-Net segmentation network employed in this project reaching a "bottleneck" in terms of performance after adding relatively few real training images. It could also be due to a combination of both; again, time did not permit any investigation of these topics.

The test performance using synthetic images exclusively is very poor, and as mentioned in section \ref{results:segmentation}, the improvements in validation error (calculated over real data) when training U-Net on this data were few and far between. This shows (almost by definition, in this context) that there is still a very large difference between the real and synthetic datasets. This may be caused by the relatively large proportion of noisy and corrupted segmentation masks in the datasets generated by the GAN, which might be polluting the synthetic datasets and overall reducing their statistical similarity to the real ones. If this is the case, it might be beneficial to apply a more sophisticated method of image preprocessing than the one employed in this project, in an attempt to clean the synthetic dataset from these particular images. For instance, one could perhaps "reuse" the trained discriminator $D$ of PGAN, by feeding it all of the images that have been generated and discarding the images that are mapped to a value below a specific threshold. One could also employ a more direct method of data cleaning, by giving a medical expert the task of going through the synthetic masks and discarding the ones that they judge unrealistic. This would obviously require human input and would somewhat defeat the purpose of automation, but it could still prove to be a big improvement in terms of effort when compared to the process of gathering patient data and annotating it by hand, if the quality of the final resulting dataset is high enough to justify this process.

It is worth mentioning that mainstream GAN architectures such as PGAN and the GAN methodology in general are not necessarily designed with discrete data in mind; the standard GAN framework relies on generated outputs being fully differentiable with respect to the generator parameters, and it is thus not adapted to discrete data such as segmentation masks (which in this project consist of arrays with integer values in [0, 6]). Some work has been done to re-purpose or adapt GANs to work more naturally with discrete data, e.g. by Devon et al. in \cite{BGAN}. Some effort was dedicated in the beginning of this project towards implementing and/or using the results of this paper, but this approach proved complicated and time-consuming, and it was ultimately abandoned in favor of a more "tried and tested" GAN framework; namely, PGAN.

The application of traditional methods of data augmentation (e.g. scaling, rotations, etc.) was not explored in this project. It would be of interest to evaluate if the performance gains that have been demonstrated in this project differ from those that are achieved when applying standard methods of augmentation. It would also be interesting to observe the difference in image quality of the generated segmentation masks, when their training data have undergone augmentation of this kind before being trained by PGAN. Lastly, it is possible that a third kind of data augmentation could be achieved ''for free'' (with respect to both segmentation and GAN training) by slicing each brain volume along different axes than just the axial, and mixing the results.

Some methodological concerns arose during the course of this project. First, the calculation of class weights and the calculation of the scaling and normalization parameters have been performed over all of the given training data in each different instance, \textit{including} the synthetic data. These values were used to transform training, validation and test data, as well as to calculate new class weights for each training distance. However, since the synthetic dataset evidently differs significantly from the real one, concerns arose that this would affect the accuracy of the validation and testing (which only used real data), as well as the loss metric (indirectly via the class weights). Empirically, it was found that the normalization parameters were not significantly affected by the inclusion of synthetic images, but that the class weights differed somewhat. With this in mind, it was considered to calculate the values for the training and test/validation separately, by using both real and synthetic training images in the former case, and only real images in the latter. It was also considered to create a validation metric separate from the training loss, as a weighted Dice loss function using weights calculated with real training data exclusively -- this function would not affect the training in any way, but would provide a means with which to monitor the validation and test results without being affected by the given synthetic dataset. Ultimately, these approaches were abandoned in favor of the simpler method of including all of the training data (real and synthetic) in all calculations, since transforming the training and test/validation data in different manners created the risk of introducing numerical bias when introducing those datasets to the trained networks.

Another concern arose during the creation of the ''reduced'' dataset consisting of only one fifth of the training images. Initially, the dataset split was performed by separating 20 \% of the \textit{brain volumes} themselves, and subsequently slicing and shuffling them, instead of first slicing and shuffling and \textit{then} separating 20 \% of the resulting slices (which is the approach that was used in the end). The former method led to very poor convergence when training the segmentation networks; the training error underwent a natural descent, but the validation error plateaued very quickly. In other words, the network experienced extreme overfitting, which did not happen when slices from all of the 210 brains in the dataset was used in the training set. This is likely a consequence of the fact that several different MR scanners have been used to collect the BraTS dataset. Different MR scanners produce different pixel intensities and have different noise properties. Therefore it was not possible (or at least not practical) to avoid mixing slices across the \textit{entire} dataset (and to create a ''true'' reduction of training data) while using the same validation and test set between each instance.

Lastly, the test results presented in \ref{table:dice} have not been repeated through cross-validation. Thus, these results (which have been produced by simply calculating the error over the entire test set each time) do not reflect the uncertainty of the different combinations of real and synthetic data. Unfortunately, since most experiments required days of training, time did not permit the proper application of cross-validation when generating test results.

\section*{Acknowledgements}

The authors thank Marco Domenico Cirillo for providing the U-Net code. Anders Eklund was supported by LiU Cancer, VINNOVA Analytic Imaging Diagnostics Arena (AIDA), and the ITEA3 / VINNOVA funded project Intelligence based iMprovement of Personalized treatment And Clinical workflow supporT (IMPACT).


\clearpage
\bibliographystyle{IEEEbib}
\bibliography{references}

\end{document}